\documentclass{article}

\usepackage{microtype}
\usepackage{graphicx}
\usepackage{subcaption}
\usepackage{booktabs} % for professional tables

\usepackage{hyperref}

\usepackage[preprint]{icml2026}

\usepackage{amsmath}
\usepackage{amssymb}
\usepackage{mathtools}
\usepackage{amsthm}

\usepackage{algorithm}

\usepackage{anyfontsize}

\usepackage{bm}
\usepackage{latexsym}
\usepackage{color}
\usepackage{epsfig}
\usepackage{xspace}
\usepackage[table]{xcolor} 
\usepackage[english]{babel}
\usepackage{array}
\usepackage{multirow}
\usepackage{makecell}
\newcolumntype{C}[1]{>{\centering\arraybackslash}p{#1}}
\newcolumntype{L}[1]{>{\raggedleft\arraybackslash}p{#1}}

\newcommand{\eat}[1]{}

\newcommand{\ie}{\emph{i.e.,}\xspace}
\newcommand{\eg}{\emph{e.g.,}\xspace}

\newcommand{\wa}[1]{\textcolor{black}{{#1}}}

\newcommand{\stab}{\rule{0pt}{2pt}\\[-2.5ex]}

\newcommand{\stitle}[1]{\vspace{0ex}\noindent{\bf #1}}

\newcommand{\mn}[1]{\emph{HyperNS}}
\usepackage[capitalize,noabbrev]{cleveref}

\theoremstyle{plain}

\theoremstyle{definition}

\theoremstyle{remark}

\usepackage[textsize=tiny]{todonotes}

\icmltitlerunning{Submission and Formatting Instructions for ICML 2026}

\begin{document}

\twocolumn[
  \icmltitle{Empowering Targeted Neighborhood Search via Hyper Tour \\for Large-Scale TSP}



  \icmlsetsymbol{equal}{*}

  \begin{icmlauthorlist}
    \icmlauthor{Tongkai Lu}{bh}
    \icmlauthor{Shuai Ma}{bh}
    \icmlauthor{Chongyang Tao}{bh}
  \end{icmlauthorlist}

  \icmlaffiliation{bh}{SKLSDE Lab, Beihang University, Beijing, China}

  \icmlcorrespondingauthor{Shuai Ma}{shuaima@buaa.edu.cn}

  \icmlkeywords{Machine Learning, ICML}

  \vskip 0.3in
]



\printAffiliationsAndNotice{}  % no special notice (required even if empty)

\begin{abstract}
  Traveling Salesman Problem (TSP) is a classic NP-hard problem that has garnered significant attention from both academia and industry. While neural-based methods have shown promise for solving TSPs, they still face challenges in scaling to larger instances, particularly in memory constraints associated with global heatmaps, edge weights, or access matrices, as well as in generating high-quality initial solutions and insufficient global guidance for efficiently navigating vast search spaces. To address these challenges, we propose a \underline{\textbf{Hyper}} Tour Guided \underline{\textbf{N}}eighborhood \underline{\textbf{S}}earch (HyperNS) method for large-scale TSP instances. Inspired by the ``clustering first, route second" strategy, our approach initially divides the TSP instance into clusters using a sparse heatmap graph and abstracts them as supernodes, followed by the generation of a hyper tour to guide both the initialization and optimization processes. This method reduces the search space by focusing on edges relevant to the hyper tour, leading to more efficient and effective optimization. Experimental results on both synthetic and real-world datasets demonstrate that our approach outperforms existing neural-based methods, particularly in handling larger-scale instances, offering a significant reduction in the gap to the optimal solution. 
\end{abstract}

\section{Introduction}\label{sec-intro}
The traveling salesman problem (TSP) is a well-known NP-hard combinatorial optimization problem. Given a list of cities (or vertices) and the distances between each pair of cities, TSP aims to find the shortest possible route that visits each city (or vertex) exactly once and returns to the origin city. 
Due to its complexity and broad application value in areas such as content delivery~\cite{lee2022economics}, robot routing~\cite{macharet2018survey}, biology~\cite{lenstra1975some}, circuit design~\cite{ducomman2016alternative}, Web crawling~\cite{tripathi2021age} and service scheduling~\cite{kiraly2015redesign}, the TSP has consistently drawn significant attention from both industry and research communities~\cite{li2024fast,grinsztajn2023winner,chalumeau2024combinatorial,luo2024neural,kwon2020pomo,jin2023pointerformer,drakulic2023bq,lyu2024scalable,zhou2023towards,son2023meta,li2025unify,wang2024asp,xiao2024distilling,li2023distribution,min2024unsupervised,ma2023learning,ye2023deepaco,wang2025efficient,li2025unify}.

Recent advances in TSP research have increasingly focused on addressing the significant challenges of large-scale instances, where the primary difficulties lie in high memory consumption and the exponentially growing search space as the problem size increases. 
Traditional methods for large-scale TSPs, such as LKH3~\cite{helsgaun2017extension}, typically employ heuristic guidance to identify promising candidate vertices during the search, thereby reducing the search space and memory consumption. However, they rely heavily on numerous expert-crafted rules and iterative searches, making them inflexible and time-consuming, especially for large-scale TSP instances.
More recently, with the introduction of neural networks into TSP solving~\cite{vinyals2015pointer}, researchers have begun exploring or combining various neural models with traditional methods to enhance TSP performance on large-scale TSPs, which can be roughly divided into heatmap-guided\footnote{Heatmaps are typically generated by GCNs, where each element represents the probability of an edge being part of the optimal TSP solution.} methods~\cite{fu2021generalize,qiu2022dimes,sun2023difusco,xia2024position,luoboosting} and decomposition-based methods~\cite{fang2024invit,ye2024glop,cheng2023select,zhou2025dualopt,zhengudc}.

While these methods have been proposed for tackling large-scale TSP, significant challenges remain for both paradigms.
First, heatmap-based methods, despite providing global guidance, inevitably incur prohibitive memory costs, as they require storing multiple $\mathbb{R}^{n\times n}$ matrices (e.g., heatmaps, edge weights, or access matrices), which severely limits their scalability to very large instances.
Second, decomposition based methods alleviate memory pressure by optimizing smaller subproblems, but the refinement of subtours is typically carried out in isolation, without explicit global coordination, making the resulting solutions prone to suboptimal global structures.
Finally, owing to the NP-hardness of TSP and the scarcity of effective heuristic guidance, both paradigms generally rely on simplistic initialization strategies such as random, greedy, or beam search. This neglect of high-quality initial solutions constitutes a critical bottleneck, often leading to slow and unstable convergence in subsequent optimization.

To address these challenges, we propose a novel framework that integrates heatmap-guided and decomposition-based approaches while mitigating their respective limitations for efficiently solving large-scale TSPs. To this end, we introduce the \underline{\textbf{Hyper}} Tour Guided \underline{\textbf{N}}eighborhood \underline{\textbf{S}}earch framework (\mn~), which systematically addresses memory bottlenecks, lack of global guidance, and poor initialization, enabling efficient routing in large-scale instances.
Motivated by the idea of ``\emph{clustering first, route second}"~\cite{gillett1974heuristic}, our method first constructs a sparse heatmap graph using a sub-graph sampling-then-aggregate paradigm, significantly reducing the memory footprint compared with full $\mathbb{R}^{n\times n}$ representations while retaining crucial edge information.
Large-scale instances are then partitioned and abstracted into supernodes by iteratively deleting bridge edges, and a hyper tour is generated over these supernodes by solving a reduced problem. This hyper tour serves as a high-level abstraction of the global path, providing consistent guidance for both initialization and subsequent optimization.
Guided by the hyper tour, we perform tour initialization to generate high-quality initial solutions, followed by a targeted neighborhood search that iteratively refines inter-supernode connections and reorganizes intra-supernode structures, focusing exclusively on relevant edges and their neighbors.
By strategically combining sparse representation, hyper tour guidance, and targeted neighborhood search, \mn~ narrows the search space, preserves global coherence, mitigates memory bottlenecks, and guides the optimization process toward faster convergence and higher-quality solutions.
Experimental results on both generated and real-world datasets, spanning various scales of TSP instances, demonstrate the effectiveness of our method compared with existing neural methods.
The main contributions of this work are as follows:

\stab(1) We propose an efficient routing framework for large-scale TSPs based on a sparse heatmap graph and a supernode-based hyper tour. The hyper tour acts as a high-level abstraction of the global path, providing global guidance that facilitates both the subsequent tour initialization and the iterative optimization of the overall solution.

\stab(2) We design a hyper-tour guided tour initialization procedure, which could generate high-quality initial solutions, thereby improving convergence speed and leading to better overall results.

\stab(3) We introduce a hyper-tour guided targeted neighborhood search, which iteratively refines the tour by focusing on targeted edges and their neighbors corresponding to the hyper tour, effectively narrowing the search space while optimizing both inter-supernode connections and intra-supernode structures.

\stab(4)  Our \mn~ can obtain optimal or near-optimal solutions within a reasonable time across uniformly generated datasets, real-world benchmarks, and non-uniformly distributed instances, clearly outperforming all existing neural-based algorithms. 
Our method can solve real-world instances up to 71,009 cites with a gap of 3.68\%.

\section{Related Work}\label{sec-related}
Early neural methods, typically limited to instances with up to 1,000 vertices, can be divided into constructive-based methods~\cite{vinyals2015pointer,bello2016neural,deudon2018learning,kool2018attention,khalil2017learning,li2024fast,grinsztajn2023winner,chalumeau2024combinatorial,luo2024neural,kwon2020pomo,jin2023pointerformer,drakulic2023bq,lyu2024scalable,jacobs21reinforcement,zhang2022learning,jiang2022learning,bi2022learning,zhou2023towards,son2023meta,li2025unify,wang2024asp}, which generate tours from scratch, and improvement-based methods~\cite{joshi2019efficient,xiao2024distilling,li2023distribution,min2024unsupervised,chen2019learning, wu2021learning,lu2019learning,hudson2022graph,yao22reversible,ma2023learning,ye2023deepaco,wang2025efficient,li2025unify}, which first generate a solution and then iteratively refine it under neural guidance toward better solutions. 
Most neural methods for large-scale TSP instances follow the improvement-based paradigm, with one exception INViT~\cite{fang2024invit}, which suffers from much lower performance. 
In this section, we focus on improvement-based methods for large-scale TSPs, which can be broadly divided into heatmap-guided methods and decomposition-based methods depending on how neural guidance is applied, while other methods are reviewed in Appendix~\ref{rel}.

Heatmap-guided methods provide global guidance by predicting a probability distribution over all edges. \textnormal{AttGCRN+MCTS\allowbreak}~\cite{fu2021generalize} extends the Att-GCRN (graph convolutional residual network with attention mechanism) model to generate heatmaps and uses MCTS for optimization, handling up to 10,000 vertices. 
DIMES~\cite{qiu2022dimes}, DIFUSCO~\cite{sun2023difusco} and SoftDist~\cite{xia2024position} improve upon this with meta-learning, diffusion models and distance based heatmap generation, respectively, but struggle with generalization to larger instances (>10,000) due to MCTS's memory constraints. % for instances >10,000 vertices

Decomposition-based methods reduce the problem scale by dividing a large instance into one or multiple subproblems, which are solved independently, thereby alleviating memory issues. Representative examples include GLOP~\cite{ye2024glop}, Select\&Optimize~\cite{cheng2023select}, DualOpt~\cite{zhou2025dualopt}, UDC~\cite{zhengudc}, and PRC~\cite{luoboosting}.
While these approaches are more scalable, their lack of global guidance makes them prone to inefficient search and local optima.
Different from these approaches, DRHG~\cite{li2025destroy} adopts a destroy-and-repair framework, which iteratively removes certain path nodes and compresses the remaining contiguous edge segments into hyper-edges, thereby significantly reducing the problem-solving scale. However, its destroy phase relies on predefined rules without a well-justified rationale and lacks explicit global guidance, leading to excessive iterations that slow down problem solving. As a result, it is prone to getting trapped in local optima and performs even worse than earlier methods on large-scale instances. In contrast, our method introduces a higher-level hyper tour abstraction that serves as a proxy for the global path, providing consistent global guidance for both initialization and iterative refinement, thereby overcoming DRHG’s local bias.

Different from most existing neural methods, we design an efficient routing method for large-scale TSP instances, capable of handling up to 70K vertices, by leveraging a cluster-based hyper tour as global guidance. 
Unlike heatmap-based approaches that rely on storing multiple large matrices, our method uses sparse heatmap graph generation combined with hyper tour-guided targeted neighborhood search, which significantly reduces memory usage while maintaining strong performance on large-scale TSPs.
Compared with divide-and-conquer based methods (\eg GLOP~\cite{ye2024glop}, Select\&Optimize~\cite{cheng2023select}, DualOpt~\cite{zhou2025dualopt}, UDC~\cite{zhengudc}, and PRC~\cite{luoboosting}), which refine subtours independently without global coordination, \mn~ constructs a global hyper tour that guides both initialization and optimization.
In particular, our initialization could speed up convergence and improve overall results, while targeted neighborhood search enforces global consistency by reorganizing connections between clusters and refining local structures. This synergy among improved initialization, global abstraction, and local refinement clearly distinguishes \mn~ from existing methods and underpins its superior performance.
\section{Methods}\label{sec-methods}
\subsection{Problem Formulation} 
We address the two-dimensional Euclidean TSP problem, which is formulated as an undirected graph $\mathbb{G(V, E)}$, where $\mathbb{V} = \{v_1,...,v_n\}$ is the set of $n$ distinct vertices (each vertex corresponds to a city) in the two-dimensional plane and $\mathbb{E}$ donates the set of edges. The goal is to find a permutation $\pi = (\pi_1, \pi_2,...,\pi_n)$, that forms a tour, visiting each vertex once and returning to the start, with the objective to minimize the total length $c(\pi)$, where $d_{i,j}$ is the Euclidean distance of two vertices $v_i$, $v_j$. 
\begin{equation}
    \setlength\abovedisplayskip{2pt}%shrink space
\setlength\belowdisplayskip{2pt}
    c(\pi) = \sum\limits_{i=1,\dots,n-1} d_{\pi_i, \pi_{i + 1}} + d_{\pi_n, \pi_1}
\end{equation}

\begin{figure*}[t]
 \centering
  \vspace{-2mm}
 \includegraphics[scale=.24]{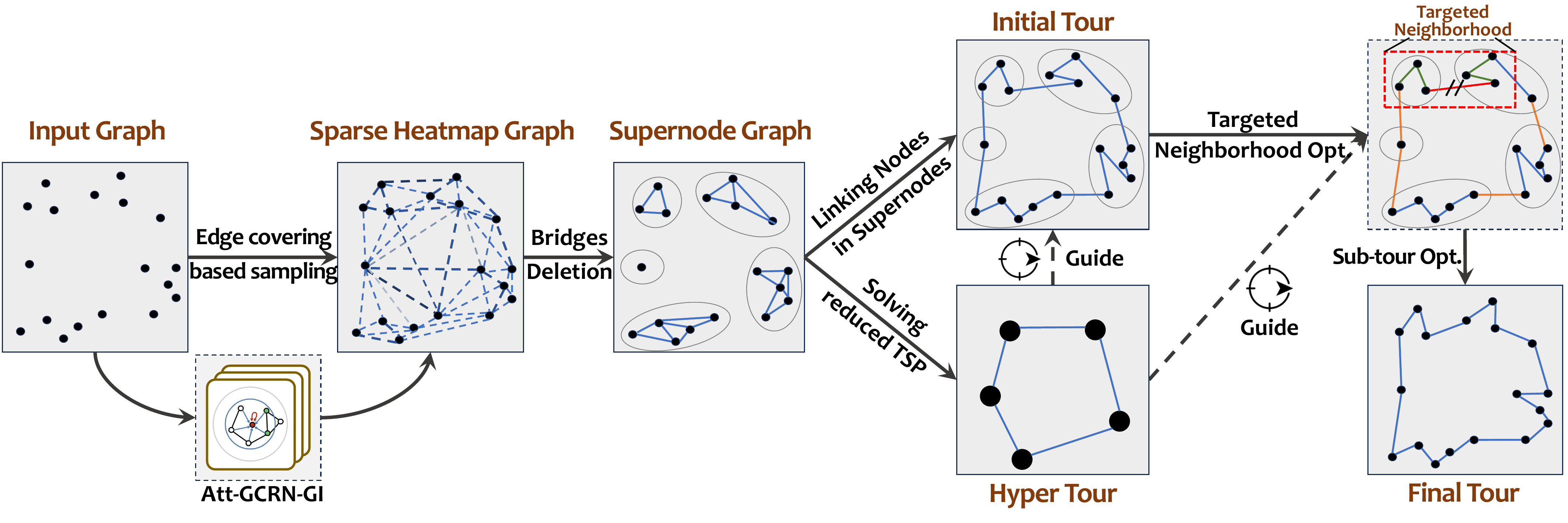}
 \vspace{-4mm}
 \caption{Overview of our \mn~ framework. Blue edges indicate normal connections, brown edges denote worth-deletion edges that link supernodes. The red edge marks the currently selected worth-deletion edge and its neighboring edges are shown in green. The area enclosed by the red dashed box in the upper-right corner illustrates the targeted neighborhood, which comprises the selected worth-deletion edge and its neighboring edges.}
  \vspace{-4mm}
 \label{framework}
\end{figure*}

\subsection{Model Overview }
In this section, we briefly introduce our \mn~ for efficient routing using hyper tours for larger-scale TSPs. Fig. \ref{framework} provides an overview of our \mn~.
First, 
considering that heatmap can help identify local structures by highlighting high-probability edges, we enhance {Att-GCRN} by incorporating geometric information, generate sub-heatmaps via edge-covering-based subgraph sampling, and merge them into a \emph{sparse heatmap graph} by selecting top-k elements.
We then partition this graph into clusters, each abstracted as a supernode, and generate a hyper tour by constructing a reduced TSP over these supernodes and solving it using Att-GCRN-GI and LK search~\cite{DBLP:journals/ior/LinK73} \footnote{The Lin–Kernighan (LK) search is currently one of the most efficient and effective algorithms for small-scale TSP instances. It is widely used as the basic search unit in methods such as LKH and MCTS due to its strong performance. \wa{For more details, please refer to Appendix~\ref{sup-LK}.}}.
The hyper tour serves as an abstraction of the global TSP solution at a higher level,
providing consistent guidance for both initialization and subsequent optimization.
Using the hyper tour as a guide, we propose a tour initialization procedure, followed by a targeted neighborhood search process. 
By restricting the optimization to hyper-tour-related edges, our method reduces the search space and progressively refines both supernode connections and intra-supernode links, ultimately yielding better solutions.

\subsection{Sparse Heatmap Graph Construction}
Considering that the heatmap generated by neural networks like GCN~\cite{khalil2017learning} and Att-GCRN~\cite{joshi2019efficient} is adopted by recent methods for large-scale TSPs, we aim to use the heatmaps as guiding signals to create a hyper tour at higher level for subsequent optimization. 
However, the current sub-graph sampling-then-aggregation paradigm~\cite{fu2021generalize,qiu2022dimes,sun2023difusco} has three issues, including missing geometry information like distances and triangle areas among vertices, random sampling without considering the coverage of uncovered vertices and the evaluation of all edges, and high memory usage for using full heatmaps. These issues not only hinder the model's ability to generalize to larger-scale problems but also reduce its solving efficiency and effectiveness.

To address these challenges, we first propose a self-supervised pretraining task inspired by~\cite{zhou2023improving} to capture geometric information like edge lengths and triangle areas and enhance the vertex representations of Att-GCRN, which is referred to as Att-GCRN-GI.
Second, we introduce an edge-covering-oriented graph sampling method that models graph sampling as a set-covering problem~\cite{young2008greedy}, aiming to minimize the number of subgraphs and sampling time while ensuring comprehensive edge coverage and then get a sparse heatmap graph directly from the sub-heatmaps.
We next detail the construction of the sparse heatmap graph with Att-GCRN-GI in the following, while leaving the architectural details of Att-GCRN-GI to Appendix~\ref{attgcrn}, as it is not the primary focus of this work.

Specifically, for each vertex $v_i$ of $\mathbb{G}$, we first get its corresponding subgraph $\mathbb{G}_i = (\mathcal{N}(v_i,p), \mathcal{E}(\mathcal{N}(v_i,p)))$, where $\mathcal{N}(v_i,p)$ is to get $v_i$'s $p$ nearest neighbors and $\mathcal{E}(S) = \{(u,v) \in \mathbb{E}| u,v\in S\}$. We regard $C_i = \mathcal{E}(\mathcal{N}(v_i,p))$ as a candidate set and can obtain subgraphs by identifying candidate sets that cover all edges for $n$ sets, thereby addressing a set-covering problem.
Next, we apply a greedy algorithm~\cite{young2008greedy} to solve it, which iteratively selects the set containing the largest number of uncovered edges at each stage.

To further optimize the process, we employ an adaptive grid partitioning method to divide the TSP instance into a series of grids, each containing no more than $\gamma$ vertices. We then randomly select a vertex from each grid to obtain a subgraph and candidate set. Following this, we generate sub-heatmaps for each subgraph with Att-GCRN-GI. 
Finally, we merge these sub-heatmaps by averaging the edge values across them and selecting the top-$k$ elements with the highest heatmap values for each vertex, resulting in a sparse heatmap graph $G_{\texttt{top-k}}$, which contains $k \times n$ edges, where $k$ is relatively small compared with the size of the instance (e.g., $k=2$ \footnote{We also tried larger $k$, which did not yield noticeable performance gains but increased memory consumption.} in our experiments).

\subsection{Hyper Tour Generation and  Initialization}
Motivated by the idea of ``clustering first, route second"~\cite{gillett1974heuristic}, we propose addressing large-scale TSP instances by dividing them into clusters based on sparse heatmap graphs, and then pursuing a hyper tour among these clusters. 
Complex optimal solutions for large-scale TSP instances often exhibit smaller, well-organized local structures, such as subloops or clusters. These local structures are crucial to the overall solution, as they capture the optimal arrangement and connections of adjacent or related elements.
Heatmaps can help identify these local structures by highlighting edges with high local probabilities, facilitating the grouping of vertices within the same sub-tour into a cluster. 
Each cluster is then abstracted into a supernode, a single vertex representing all vertices within the cluster—allowing us to form a smaller TSP instance while preserving the essential structure of the original graph. We solve this reduced problem and refer to its solution as the ``hyper tour", 
which determines how the supernodes are connected and serves as a guide for generating and optimizing the initial solution. 

\stitle{Hyper Tour Generation.} As outlined in Algorithm \ref{alg:hypertour}, we first cluster the vertices by iteratively deleting bridge edges and then abstract each cluster into a supernode representing the vertices inside it. 
More specifically, it first gets all connected subgraphs of $G_\texttt{top-k}$ (Line 3). 
Then for each connected subgraph, it gets and deletes its bridge edges in $G_\texttt{top-k}$ (Line 7-8).
Repeat the above steps until all connected subgraphs of $G_\texttt{top-k}$ have no bridge edges (Line 4-5).
Third, we construct a reduced TSP problem defined over the supernodes (Line 9) and solve it with Att-GCRN-GI and LK search, which provides a simple yet effective way to handle such reduced instances efficiently (Line 10).
We take the average coordinates of nodes within the cluster as the new supernode's coordinates.
Finally, we get the hyper tour and the supernode list (Line 11).
\begin{algorithm}[!t]
    \setlength\abovedisplayskip{2pt}%shrink space
    \setlength\belowdisplayskip{2pt}
    \caption{Hyper Tour Generation}
    \label{alg:hypertour}
    \small
    \begin{algorithmic}[1]
        
        \REQUIRE the sparse heatmap graph $G_\texttt{top-k}$.
          \ENSURE Sketch tour $T_s$ and cluster list $\texttt{Cluster}$.
          %  \Procedure {get\_non\_redundant\_rules\_1}{$S$}
          % \eat{\State Let $G_{top-2}$.node=$V$; $G_{top-2}$.edge=[];
          % \FOR{$i \in V$}
          % \STATE $j,k$ = Get\_top2\_index($P[i,:]$); 
          % \STATE $G_{top-2}$.edge.append(($i,j$));$G_{top-2}$.edge.append(($i,k$));
          % \ENDFOR}
          \STATE $l$ = len(get\_connected\_subgraph($G_\texttt{top-k}$));
          \WHILE{True}
          \STATE $\texttt{Cluster}$ = get\_connected\_subgraph($G_\texttt{top-k}$);
          \IF{$l$ == len($\texttt{Cluster}$)} 
          \STATE break;
          \ENDIF
          \STATE $l$ = len($\texttt{Cluster}$);
          \FOR{subgraph $ \in $ $\texttt{Cluster}$} 
          \STATE $G_\texttt{top-k}$ = delete\_bridges(subgraph, $G_\texttt{top-k}$);
          \ENDFOR
          \ENDWHILE
          % \State $C$ = subgraph\_list;
          \STATE $G_{s}$ = get\_small\_graph($\texttt{Cluster}$);
          \STATE $T_s$ = solve\_small\_tsp($G_{s}$);
        \STATE  \textbf{return} $T_s$, $\texttt{Cluster}$.
        
    \end{algorithmic}
\end{algorithm}

The absence of bridge edges ensures that every pair of vertices within the supernode is connected by at least two paths. 
{This means that vertices densely connected in the small instance $G_s$ are likely to be strongly connected in the final solution.} Thus, the hyper tour plays a crucial role in providing an initial solution and guiding the subsequent optimization, underscoring the importance of our hyper tour concept. 

\begin{figure*}[t]
 \centering
  \vspace{-2mm}
 \includegraphics[scale=.24]{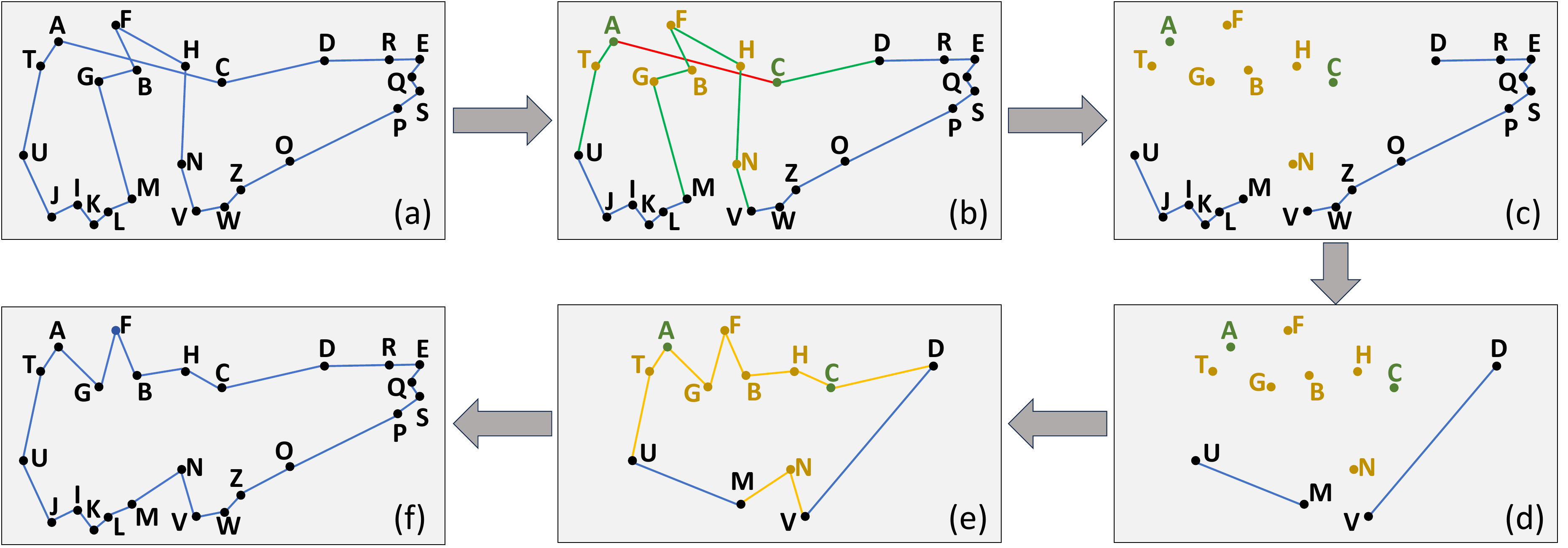}
 % \vspace{-2mm}
 \caption{Illustration of targeted neighborhood search procedure. (a) A TSP tour. (b) Select edge $(A, C)$ along with $A$ and $C$'s $3$-nearest neighbors: $\{T, G, F, B, H, N\}$ and identify all edges involving them as $\mathbb{E}_{del}$. (c) Delete all edges in $\mathbb{E}_{del}$. (d) Retain only vertices $\{U, M, V, D\}$ while adding two new edges, $(U, M)$ and $(V, D)$ to form a reduced subproblem with no more than 12 vertices. (e) Solve the subproblem via LK search, keeping fixed edges $(U, M)$ and $(V, D)$ unchanged. (f) Restore the fixed edges to their previous segments, yielding an improved version of the initial tour.}
  % \vspace{-5.5mm}
 \label{globalSearch}
\end{figure*}

\stitle{Hyper-Guided Tour Initialization.}
A well-chosen initial solution can speed up convergence and lead to better overall results. However, due to the NP-hardness of TSP and the lack of heuristic guidance, efficiently generating one for large-scale TSPs is particularly challenging, especially for large-scale TSPs. Existing neural-based methods often pay little attention to initial tour generation, typically relying on simple methods.
In our method, the supernode list and hyper tour enable the efficient generation of a higher-quality initial tour, which contributes to better final performance.

Specifically, the hyper tour could provide a rough approximation of the optimal TSP tour at a higher level.
We first connect the supernodes guided by the hyper tour, forming a ring-like structure where each supernode $\hat{C}_i$ is connected to its two neighbors $\hat{C}_{i-1}, \hat{C}_{i+1}$.
Second, we select any supernode from the ring as the starting point and traverse through neighboring supernodes in a specified direction along the ring. This traversal continues until the total number of vertices in the traversed supernodes exceeds a predefined threshold length $l_s$. Let $V_{traversed} = \bigcup_{i=0,1,..,k}\hat{C}_{start+i}$ be the the traversed supernodes, we have $|V_{traversed}| > l_s$ and $|\bigcup_{i=0,1,..,k-1}\hat{C}_{start+i}| < l_s$. %rapid local optimization
Third, we apply the LK search algorithm to connect the vertices within $V_{traversed}$ efficiently. 
During this process, we also record the edges connecting different supernodes as ``worth-deletion" edges. These edges are candidates for removal in later optimization stages to enhance the overall tour.
Finally, we designate the last traversed supernode as the new starting supernode and repeat the traversal and connection steps until we return to the original starting supernode. This approach yields an initial tour and a set of worth-deletion edges.

\subsection{Targeted Neighborhood Search} 

Since the worth-deletion edges obtained during the initialization procedure are only rough approximations of the optimal tour, some of them may be incorrectly connected. To address this, we propose a targeted neighborhood search procedure that iteratively removes these suboptimal edges and repairs the partial tour with a targeted LK search with each iteration focusing on optimizing connections among a targeted edge and its neighborhood.

Similar to MCTS, our targeted neighborhood search is considered as a Markov Decision Process (MDP), which starts from an initial state $\pi$, and iteratively applies an action $a$ to reach a new state $\pi^{new}$. However, instead of k-opt search, we adopt a new kind of action called \emph{destroy-and-repair}, which first destroys the current solution by deleting an edge with the highest score and all edges associated with its closed neighborhood, and then repairs them using the traditional LK search. This iterative process gradually reforms potentially incorrect connections between supernodes, leading to an improved solution.

1) Initialization.
The distance of each edge initializes the weight dict $W[(i,j)] = 100 \times d_{i,j}$, which records the weight of each edge in the current tour where $i<j$. 
The dict $Q$ starts with all elements set to zero, which records the duration times in which the edges remain undeleted by the "destroy" action.
The dict $O$ is initialized with worth-deletion edges set to 10,000 and others set to zero, which is to prioritize exploring longer worth deletion edges first, enhancing search efficiency.

2) Selection.
At each iteration, select the edge with the highest score in the current solution. The score $Z_{i,j} (i<j)$ of an edge $(v_i, v_j)$ consists of 3 parts:
\begin{equation}
    \setlength\abovedisplayskip{2pt}%shrink space
\setlength\belowdisplayskip{2pt}
\hspace{-2mm}Z_{i,j} = W[(i,j)] + \alpha\frac{ Q[(i,j)]}{1 + \sum_{(\bar{i},\bar{j})\in \pi} Q[(\bar{i},\bar{j})]} + O[(i,j)]
% \vspace{-1mm}
\end{equation}
In this formula, the first part emphasizes the importance of the edges with high $W[(i,j)]$ values, while the second part prefers the rarely affected edges. $\alpha$ is a parameter to balance intensification and diversification, and the term ”+1” is used to avoid a zero denominator. The third part is designed to explore worth deletion edges first and exclude edges that have been previously selected.

3) \emph{Destroy and repair}. Fig. \ref{globalSearch} details our \emph{destroy and repair} process. Starting with a TSP tour (Fig.~\ref{globalSearch} (a)), we first get an edge set $\mathbb{E}_{del}$ by selecting an edge $(v_{\pi_i}, v_{\pi_{i+1}})$ with the highest score, and taking the two endpoints of the edge and the $m$-nearest neighbors of these endpoints  (Fig.~\ref{globalSearch} (b)). 
\begin{equation}
    \setlength\abovedisplayskip{4pt}%shrink space
\setlength\belowdisplayskip{4pt}
\mathbb{E}_{del}\! = \!\{(v_i, v_j)\! \in \!T_\pi| v_i \ \mathrm{or}\  v_j \in \mathcal{N}(v_{\pi_i},\! m) \bigcap \mathcal{N}(v_{\pi_{i+1}},\! m)\}
% \nonumber
\end{equation}
where $ T_\pi$ is the set of edges corresponding to the permutation $\pi$, and $\mathcal{N}$ is the function used to retrieve the neighbors.
Next, we delete all edges in $\mathbb{E}_{del}$ (Fig.~\ref{globalSearch} (c)) and merge the left segments by preserving the two endpoints of each segment, connecting them to an edge that remains fixed during subsequent optimization. This yields a reduced subproblem with no more than $4m$ vertices (Fig.~\ref{globalSearch} (d)). 
In the third step, we solve the subproblem via LK search, keeping fixed edges unchanged (Fig.~\ref{globalSearch} (e)).  Finally, we restore the fixed edges to their previous segments, yielding an improved initial tour (Fig.~\ref{globalSearch} (f)). The iteration of our targeted neighborhood search stops when the change is less than 0.01\% for 10 consecutive iterations.

4) Updating. The three dictionaries are updated as follows: First, for the selected edge, set its value in $O$ to negative infinity. Second, for newly added edges, initialize their values in all three dictionaries as previously described. Third, for deleted edges, remove their corresponding values from the three dictionaries. Fourth, for unaffected edges, increment their $Q$ value by 1. Finally, for edges that were deleted during the \emph{destroy} phase but reintroduced during the \emph{repair} phase, \ie those that survive the destroy process, we update their weight as follows:
\begin{equation}
    \setlength\abovedisplayskip{2pt}%shrink space
\setlength\belowdisplayskip{2pt}
    \begin{aligned}        
        W[(i,j)]  = W[(i,j)] \times (1 - [&\mathrm{exp}\frac{L(\pi) - L(\pi^{new})}{L(\pi)} - 1]) \\
        Q[(i,j)] &= 0
    \end{aligned}
    % \nonumber 
\end{equation}
where $L$ is the tour length. 
It is worth noting that we only update the weight when an edge survives in the destroy phase, which reduces the probability of selecting good edges. Additionally, we only record the features of the edges present in the current solution. This significantly reduces the storage space compared with MCTS and {LKH3} ($O(n^2) $ to $O(n)$), enabling our method to be applied to larger-scale data. 
Our method also uses segment merging—only retaining the endpoints of each segment—to break down problems of any size into manageable subproblems, while ensuring that the optimization problem remains consistent before and after the transformation. {Besides, our score function purely relies on edge distances and avoids complex computations like $\alpha$-measure of LKH3.}

\stitle{Sub-tour Optimization.}
Since our targeted neighborhood search may disrupt connections within supernodes, we further propose a sub-tour optimization to reorganize these connections within a supernode and its neighboring supernodes on a small scale by refining sub-tours of fixed length. 
Formally, starting from a selected vertex $v_0$, we divide the current tour into several sub-tours $T_{sub}(i)$, each containing $l_s$ vertices, following a given direction. Each neighboring sub-tour shares a common vertex $v_{i\cdot l_s}$ with the next, ensuring continuity among supernodes.
\begin{equation}
    \setlength\abovedisplayskip{2pt}%shrink space
\setlength\belowdisplayskip{2pt}
    \hspace{-1.5mm}T_{sub}(i) = \{v_{(i-1)\cdot l_s + 1},v_{(i-1)\cdot l_s + 2}, ..., v_{\mathrm{min}(i\cdot l_S, n)}\}
\end{equation}

The sub-tours are then optimized via LK search efficiently, which minimizes their total length while keeping endpoints fixed. The process is repeated  $I_3$ times by selecting different starting vertices.

\stitle{Space complexity.}
For sparse heatmap graph construction, the subgraph and the sub-heatmaps take $O(tp^2)$ space, and the sparse graph $G_\texttt{top-k}$ takes $O(kn)$ space.
For the hyper tour generation, the supernodes and the hyper tour take in total less than $O(n)$ space.
For tour initialization, the initial tour itself takes $O(n)$ space.
The small graph for targeted neighborhood search and subtour for subtour optimization takes $O(m)$ and $O(l_s)$ space, which are all less than $n$.
Therefore, our \mn~ method in total takes $O(n)$ space. For the time complexity, please refer to appendix~\ref{complexity}.
\begin{table*}[t!]
	\centering
     \small
	 \setlength{\tabcolsep}{2pt}
	\caption{Performance on generated datasets~\cite{luoboosting}. ``N/A" indicates failure to solve due to memory constraints.}
% \vspace{-2mm}
		\resizebox{\textwidth}{!}{
        \begin{tabular}{c|c|c|c|c|c|c|c|c|c|c}
			\Xhline{2\arrayrulewidth}
			\multirow{2}{*}{Methods}&\multicolumn{2}{c|}{TSP1K}&\multicolumn{2}{c|}{TSP5K}&\multicolumn{2}{c|}{TSP10K}&\multicolumn{2}{c|}{TSP20K}&\multicolumn{2}{c}{TSP50K}\\
			\cline{2-11}
			& Length (Gap) &Time & Length (Gap) &Time&Length (Gap) &Time  &Length (Gap) &Time &Length (Gap) &Time \\
			\hline
			Concorde~\cite{cook2011traveling}&23.12 (0\%)&5.78h&N/A&N/A&N/A&N/A&N/A&N/A&N/A&N/A\\
			LKH3~\cite{helsgaun2017extension}
			&23.12 (0\%)&1.69h&50.90 (0\%)&2.66h&71.77 (0\%)&2.52h&101.31 (0\%)&7.33h&159.93 (0\%)&50.3h\\
			\hline   
			AM-greedy~\cite{kool2018attention} &33.55 (45.1\%)&3.45m&96.74 (90.1\%)&4.03m&153.42 (113.8\%)&2.13m&244.20 (141.0\%)&5.13m&N/A&N/A\\
			GCN-BS~\cite{joshi2019efficient}&51.1 (121.0\%)&3.23h&N/A&N/A&N/A&N/A&N/A&N/A&N/A&N/A\\
			\scalebox{0.9}[1]{AttGCRN+MCTS~\cite{fu2021generalize}} &23.86 (3.2\%)&0.17h&52.69 (3.5\%)&1.15h&74.93 (4.4\%)&1.18h&N/A&N/A&N/A&N/A\\
			DIMES~\cite{qiu2022dimes}&23.69 (2.5\%)&4.62h&52.37 (2.9\%)&3.47h&74.06 (3.2\%)&3.57h&N/A&N/A&N/A&N/A\\
			DIFUSCO~\cite{sun2023difusco}&23.39 (1.2\%)&0.41h&52.04 (2.2\%)&0.71h&73.62 (2.6\%)&0.79h&N/A&N/A&N/A&N/A\\
			SoftDist~\cite{xia2024position}&23.63 (2.2\%)&9.88m&52.25 (2.7\%)&1.03h&74.03 (3.1\%)&1.05h&N/A&N/A&N/A&N/A\\			
			SO~\cite{cheng2023select}&23.77 (2.8\%)&0.93h&52.60 (3.3\%)&1.92h&74.30 (3.5\%)&2.03h&105.03 (3.7\%)&8.92h&N/A&N/A\\
            UDC~\cite{zhengudc}&23.53(1.8\%)& 0.57h&33.26(2.5\%)&0.21h&N/A&N/A&N/A&N/A&N/A&N/A\\
			DRHG~\cite{li2025destroy}&\textbf{23.19 (0.3\%)}&2.31h&51.39 (0.9\%)&3.13h&72.85 (1.3\%)&4.26h&103.75 (2.4\%)&6.14h&167.23 (4.6\%)&9.35h\\
			INViT~\cite{fang2024invit}&24.50 (6.0\%)&0.26h&54.19 (6.5\%)&0.76h&76.12 (6.1\%)&0.73h&108.49 (7.1\%)&5.68h&171.38 (7.2\%)&16.76h\\
            PRC~\cite{luoboosting}&23.37 (1.1\%)&3.47h&51.67 (1.3\%)&6.23h&73.08 (1.8\%)  & 3.71h&103.66 (2.3\%)&5.93h&163.95 (2.5\%)&16.92h\\
			DualOpt~\cite{zhou2025dualopt}&23.31 (0.8\%)&0.31h&51.56 (1.3\%)&0.64h&72.62 (1.2\%)&0.87h&102.90 (1.6\%)&2.28h&162.81 (1.8\%)&7.83h\\
			\hline
			\mn~ &23.20 (0.3\%)&0.89h&\textbf{51.06 (0.3\%)}&0.94h&\textbf{72.44
			 (0.9\%)}&0.93h&\textbf{102.36 (1.0\%)}&2.19h&\textbf{162.45 (1.6\%)}&6.56h\\
			\Xhline{2\arrayrulewidth}
			
		\end{tabular}
        }
		
		% \vspace{-3mm}
		\label{T1}
	\end{table*}

\section{Experimental Study}
\label{sec-exp}
In this section, we evaluate the effectiveness of our \mn~ on both synthetic datasets with diverse distributions and real-world benchmarks, while leaving a dedicated study on the interaction between hyper tour quality and final solution performance to Appendix~\ref{exp:hyper}.

\subsection{Experimental Settings}
\stitle{Datasets.}
We first evaluate our approach on the standard TSP benchmarks~\cite{fu2021generalize,luoboosting,qiu2022dimes,li2025destroy,zhou2025dualopt}, uniformly generated within a unit square (128 instances with $N$=1K; 32 with $N$=5K; 16 each with $N$=10K, 20K, 50K).
Beyond standard benchmarks, we further evaluate the generalization of \mn~ on real-world and cross-distribution datasets. We use TSPLIB~\cite{reinelt1991tsplib} with diverse instances, including original TSPLIB (21 instances, 150–18,512 cities), Nationallib (18 instances, 194–71,009 cities), and VLSI (84 instances, 131–52,057 cities)—as well as cross-distribution datasets from INVIT~\cite{fang2024invit} with four node distributions (uniform, clustered, explosion, implosion), enabling a thorough evaluation of robustness and generalization.

\stitle{Evaluation Metric.} Following standard TSP benchmarks, we report the average tour length, gap to the optimal (computed via Concorde or LKH3), and total runtime for solving all instances to highlight the cumulative computational resources required. 

\begin{table*}[t!] 
	\centering
	\small
	\setlength{\tabcolsep}{1mm}
	\caption{Performance on TSPLIB~\cite{reinelt1991tsplib}.}
	\vspace{-2mm}
	% \resizebox{0.9\textwidth}{!}{  
		\begin{tabular}{c|cc|cc|cc|cc}
			\Xhline{2\arrayrulewidth}
			\multirow{2}{*}{Size} 
			& \multicolumn{2}{c|}{LKH3~\cite{helsgaun2017extension}} 
			& \multicolumn{2}{c|}{DRHG~\cite{li2025destroy}} 
			& \multicolumn{2}{c|}{DualOpt~\cite{zhou2025dualopt}} 
			& \multicolumn{2}{c}{\mn~} \\
			\cline{2-9}
			& Gap (\%) & Time 
			& Gap (\%) & Time 
			& Gap (\%) & Time 
			& Gap (\%) & Time \\
			\hline
			0–1K & 0.01 & 0.4h & 1.07 & 0.47h & 1.13 & 0.16h & 1.01 & 0.22h \\
			1K–5K & 0.02 & 11.4h & 2.31 & 10.74h & 2.52 & 3.92h & 2.02 & 4.26h \\
			5K–10K & 0.03 & 10.6h & 2.43 & 13.15h & 1.99 & 3.03h & 1.70 & 3.14h \\
			10K–20K & 0.04 & 12.4h & 2.79 & 6.83h & 2.01 & 3.56h & 1.83 & 3.35h \\
			\Xhline{2\arrayrulewidth}
		\end{tabular}
	% }
	\vspace{-1mm}
	\label{T2}
    % \vspace{-3mm}
\end{table*}

\stitle{Comparison methods.}
We compare our algorithm against a range of TSP solvers: (1) classic solvers like Concorde~\cite{cook2011traveling} and LKH3~\cite{helsgaun2017extension}{\footnote{LKH3 is a highly specialized, manually crafted solver that incorporates extensive domain knowledge and sophisticated heuristics, such as k-opt, $\alpha$-measure, partitioning and tour merging methods, iterative partial transcription, and backbone-guided search.}}, and (2) neural-based methods such as AM-greedy~\cite{kool2018attention}, GCN-BS~\cite{joshi2019efficient}, AttGCRN+MCTS~\cite{fu2021generalize}, DIMES~\cite{qiu2022dimes}, DIFUSCO~\cite{sun2023difusco}, SoftDist~\cite{xia2024position}, Select\&Optimize (SO)~\cite{cheng2023select}, UDC~\cite{zhengudc}, 
DRHG~\cite{li2025destroy}, INViT~\cite{fang2024invit}, PRC~\cite{luoboosting}, and DualOpt~\cite{zhou2025dualopt}. 
All baselines are run with default settings, except Select\&Optimize, which we reimplemented due to the unavailable source code. 
{Since our main focus is on integrating neural and traditional methods, we primarily compare our \mn~ with similar approaches and list the results of LKH3 for indicative purposes following previous neural methods.}

\stitle{Implementation details.}
Our experiments were conducted on an Intel Xeon Gold 6148 CPU@2.40GHz and an NVIDIA Tesla V100 PCIe 32GB GPU. The GPU was used for neural models, and the CPU for conventional solvers.
Att-GCRN-GI\footnote{The trained model is fixed and directly applied to TSP instances of varying sizes and distributions without further retraining.} is trained on 900K instances with $n$=$100$, following the standard configuration in prior works~\cite{joshi2019efficient,fu2021generalize}. 
The subgraph size $p$ and the number of vertices per grid $\gamma$ for sparse heatmap graph generation are set to 100 and 30, respectively, to match Att-GCRN-GI’s training graph sizes.
$\alpha$ for targeted neighborhood search is set to 1000 to maintain consistent magnitudes across different terms of the score.
Based on the experimental results in {Exp-6}, we set the nearest neighbor number $m$ to 100, sub-tour length $l_s$ to 100 and the number of iterations $I_3$ for sub-tour optimization to 2.

\begin{table*}[t!]
		\small
   \setlength{\tabcolsep}{0mm}
   \caption{Results on large-scale real-world instances~\cite{reinelt1991tsplib}. For neural methods, only DualOpt is reported, as it achieves the best results among all neural baselines (Tables~\ref{T1} \& ~\ref{T2}).}
		\begin{center}
    \resizebox{1\textwidth}{!}{
            \begin{tabular}{c|c|c|c|c|c|c|c|c|c|c|c|c|c|c|c}
        \Xhline{2\arrayrulewidth}
     Method&Instance
      & sw24978 & bbz25234 & irx28268 & fyg28534 & icx28698 & boa28924 & pbh30440 & fry33203 & bm33708 & pba38478 & rbz43748 & fht47608 & fna52057&ch71009\\
					\hline
     \multirow{2}{*}{LKH3~\cite{helsgaun2017extension}}&gap (\%)  & 0.00  & 0.03  & 0.02  & 0.01  & 0.02  & 0.02  & 0.01  & 0.03  & 0.01  & 0.04  & 0.03  & 0.04  & 0.03&0.05\\
     \cline{2-16}
     &Time & 2.5h & 2.02h & 1.77h & 2.26h & 2.08h & 2.31h & 2.25h & 2.21h & 4.75h & 3.22h & 5.12h & 5.70h & 5.28h&8.76h\\
     \hline
     \multirow{2}{*}{DualOpt~\cite{zhou2025dualopt}} &gap (\%)& 3.63&4.01&3.54 & 3.05 & 2.71 & 2.21 & 3.37 & 3.35 & 2.36 & 3.74 & 4.11 & 3.83 & 3.76 & 3.91 \\
     \cline{2-16}
     &Time &0.78h&0.83h&0.81h& 0.84h& 0.84h& 0.93h& 0.97h& 1.12h& 1.08h& 1.35h& 1.74h& 1.96h& 2.11h & 2.74h \\
     \hline
     \multirow{2}{*}{\mn}&gap (\%)  & 2.39  & 3.35  & 3.29  & 3.08  & 2.57  & 2.41  & 3.15  & 3.51  & 2.23  & 3.24  & 3.70  & 3.47  & 3.31&3.68\\
     \cline{2-16}
     &Time & 0.65h & 0.75h & 0.76h & 0.78h & 0.79h & 0.82h & 0.85h & 0.90h & 0.91h & 1.12h & 1.38h & 1.54h & 1.67h&2.13h\\
	\Xhline{2\arrayrulewidth}
				\end{tabular}}
		\end{center}
     
     \label{T3}
	\end{table*}
    
\begin{table*}[t!]
\small
\caption{Performance on non-uniform datasets~\cite{fang2024invit}.}
\resizebox{1\textwidth}{!}{
\begin{tabular}{c|cc|cc|cc|cc|cc|cc}

\Xhline{2\arrayrulewidth}
\textbf{Distribution} & \multicolumn{6}{c|}{\textbf{uniform}}                                                        & \multicolumn{6}{c}{\textbf{Clustered}}                                                      \\
\hline
\multirow{2}{*}{Methods}     & \multicolumn{2}{c|}{TSP1K} & \multicolumn{2}{c|}{TSP5K} & \multicolumn{2}{c|}{TSP10K} & \multicolumn{2}{c|}{TSP1K} & \multicolumn{2}{c|}{TSP5K} & \multicolumn{2}{c}{TSP10K} \\
\cline{2-13}
& Gap (\%)        & Time    & Gap (\%)        & Time    & Gap (\%)         & Time    & Gap (\%)        & Time    & Gap (\%)        & Time    & Gap (\%)         & Time    \\
\hline
DRHG~\cite{li2025destroy}         & 0.31            & 2.31h   & 0.88            & 3.13h   & 1.33             & 4.26h   & 0.49            & 2.33h   & 1.35            & 3.16h   & 2.14             & 4.27h   \\
INViT~\cite{fang2024invit}        & 5.99            & 0.26h   & 6.46            & 0.76h   & 6.06             & 0.73h   & 8.63            & 0.26h   & 8.57            & 0.76h   & 8.79             & 0.73h   \\
DualOpt~\cite{zhou2025dualopt}      & 0.82            & 0.31h   & 1.3             & 0.64h   & 1.18             & 0.87h   & 1.07            & 0.31h   & 1.68            & 0.67h   & 2.32             & 0.93h   \\
\hline
\mn~      & \textbf{0.3}    & 0.89h   & \textbf{0.31}   & 0.94h   & \textbf{0.93}    & 0.93h   & \textbf{0.41}   & 0.89h   & \textbf{0.55}   & 0.94h   & \textbf{1.53}    & 0.94h   \\
\Xhline{2\arrayrulewidth}
\textbf{Distribution} & \multicolumn{6}{c|}{\textbf{Explosion}}                                                      & \multicolumn{6}{c}{\textbf{Implosion}}                                                      \\
\hline
\multirow{2}{*}{Methods}   & \multicolumn{2}{c|}{TSP1K} & \multicolumn{2}{c|}{TSP5K} & \multicolumn{2}{c|}{TSP10K} & \multicolumn{2}{c|}{TSP1K} & \multicolumn{2}{c|}{TSP5K} & \multicolumn{2}{c}{TSP10K} \\
\cline{2-13}
  & Gap (\%)        & Time    & Gap (\%)        & Time    & Gap (\%)         & Time    & Gap (\%)        & Time    & Gap (\%)        & Time    & Gap (\%)         & Time    \\
\hline
DRHG~\cite{li2025destroy}         & 0.43            & 2.31h   & 1.47            & 3.14h   & 2.53             & 4.29h   & 0.41            & 2.31h   & 1.32            & 3.13h   & 2.48             & 4.26h   \\
INViT~\cite{fang2024invit}        & 8.57            & 0.26h   & 9.43            & 0.76h   & 9.05             & 0.74h   & 6.35            & 0.26h   & 7.41            & 0.76h   & 6.21             & 0.73h   \\
DualOpt~\cite{zhou2025dualopt}      & 1.01            & 0.32h   & 1.54            & 0.65h   & 1.86             & 0.91h   & 0.84            & 0.33h   & 1.47            & 0.66h   & 1.75             & 0.88h   \\
\hline
\mn~        & \textbf{0.34}   & 0.89h   & \textbf{0.39}   & 0.94h   & \textbf{1.11}    & 0.93h   & \textbf{0.34}   & 0.89h   & \textbf{0.43}   & 0.94h   & \textbf{1.21}    & 0.93h \\ 
\Xhline{2\arrayrulewidth}
\end{tabular}
}
% \vspace{-4mm}
\label{Tnon}
\end{table*}

% \vspace{-2mm}
\subsection{Experimental Results}
\stitle{\wa{Exp-1: Performance on Generated Datasets.}}
We first compare our \mn~ method with baseline methods on medium and large-scale generated datasets. The results are presented in Table \ref{T1}. Our \mn~ consistently outperforms nearly all neural-based baselines across all benchmarks with the only exception being DRHG on TSP1K and achieves the highest efficiency on larger-scale instances like TSP50K. 
The average gaps across five datasets produced by \mn~ are 0.5\%, 1.0\%, 0.3\%, 0.6\%, and 0.2\% lower than those of DualOpt, the strongest neural-based baseline capable of handling the largest-scale instances. Although DualOpt and INViT exhibit higher efficiency on smaller-scale instances compared with our \mn~, their computational time grows more rapidly with instance size, resulting in lower efficiency on larger problems.
Additionally, \mn~ offers an about $5\times$ speedup over LKH3 while maintaining competitive solution quality and much lower space complexity.

While AM-greedy is the fastest method, its efficiency comes at the cost of a significantly higher average gap.
Methods like AttGCRN+MCTS, DIMES, DIFUSCO, and SoftDist fail on TSP20K due to platform memory constraints imposed by MCTS and global heatmap. 
SO and INViT can solve larger instances by focusing on optimizing selected sub-tours of the current solution, sacrificing performance for lower memory usage.
DRHG, on the other hand, suffers from low efficiency due to excessive search iterations caused by the lack of effective global guidance. 
In contrast, our targeted neighborhood search employs segment merging to reduce instance size while preserving global tour length.  
These results underscore the superiority of our \mn~ for solving larger-scale TSPs.

\stitle{\wa{Exp-2: Performance on Real-World Datasets.}}
We further compare our \mn~ with three different methods on real-world datasets. We divide the TSPLIB data into four groups: 0-1K with 23 
instances, 1K-5K with 60 instances, 5K-10K with 16 instances, and 10K-
20K with 10 instances. As shown in Table~\ref{T2}, compared with the two neural methods, DRHG and DualOpt, the performance of \mn~ on real-world datasets remains competitive and shows minimal degradation relative to its performance on uniformly distributed generated datasets. Our method also achieves significant speedups over LKH3 on real-world datasets, being 1.82, 2.67, 3.38, and 3.70 times faster, respectively, with the efficiency advantage increasing as the instance size increases. Additionally, the gap of \mn~ from the optimal solution remains relatively stable as the scale grows, highlighting the generalization capability of our method for large-scale real-world datasets.

\stitle{Exp-3: Performance on Large-Scale Instances.} 
We also evaluate our method on 14 larger-scale instances ranging from 24,978 to 71,009 vertices from TSPLIB compared with LKH3 and DualOpt. 
As shown in Table~\ref{T3}, we outperform DualOpt on 11 out of 14 instances and achieve an average gap that is 8.83\% lower than that of DualOpt, with the highest gap under 4\% on rbz43748.
\mn~ is the fastest among all methods, running $1.20$ and $3.34$ times faster than DualOpt and LKH3, respectively.
Additionally, as the instance size increases, our \mn~ maintains a near-linear runtime growth, while DualOpt’s runtime increases much more rapidly, and LKH3 exhibits significant instability.
Notably, our \mn~ could achieve a 3.68\% gap on real-world instances with up to 71,009 cities, demonstrating competitive performance with LKH3 on very large instances while being about three times faster.

\stitle{Exp-4: Performance on Non-Uniform Instances.}\label{sec-non}
We evaluate the generalization ability of our method on cross-distribution instances with DRHG~\cite{li2025destroy}, INViT~\cite{fang2024invit} and DualOpt~\cite{zhou2025dualopt}.
The results are shown in Table~\ref{Tnon}.
Our \mn~ achieves the lowest gap among these methods on all datasets, showing its great cross-distribution generalizability. Specifically, the average gaps across different distributions produced by \mn~ are 0.59\%, 0.86\%, 0.86\%, and 0.69\% lower than those of DualOpt. Additionally, the relative gap increase from uniform distribution to other distributions for \mn~ on TSP10K is much lower than DualOpt (38.0\% v.s. 68.5\% on average). This indicates that our proposed method possesses both excellent cross-distribution and cross-size generalizability.

\begin{table*}[t!]
	\centering
	\small
    \caption{Ablation studies. The gray row is our full model. HTI and HTO refer to hyper tour-guided initialization and optimization.}
    % \vspace{-3mm}
		\begin{tabular}{c|c|c|c|c|cc|cc|cc}
			\Xhline{2\arrayrulewidth}
			\multirow{2}{*}{ID}&Neural&Graph&Initiali-&Optimi-&\multicolumn{2}{c|}{TSP5K}&\multicolumn{2}{c|}{TSP10K}&\multicolumn{2}{c}{TSP20K}\\
          
			\cline{6-11}&Model
			&Sampling&zation&zation& Gap &Time&Gap&Time  &Gap &Time \\
                \hline
                \rowcolor{gray!40} 0&Att-GCRN-GI&Edge covering&HTI&HTO&
            0.31\%&0.94h&0.93\%&0.93h&1.04\%&2.19h\\
             \hline
            1&\cellcolor{yellow!35} Att-GCRN& Edge covering&HTI&HTO&0.45\%&0.96h&1.00\%&0.95h&1.14\%&2.22h\\
            \hline
			2&Att-GCRN-GI&\cellcolor{yellow!35} Node covering&HTI&HTO&0.60\%&1.00h&1.12\%&1.03h&1.30\%&2.31h\\
			
			\hline

			3&Att-GCRN-GI&Edge covering&\cellcolor{yellow!35}Greedy&HTO&4.31\%&0.75h&6.90\%&0.76h&9.25\%&1.71h\\
                4&Att-GCRN-GI&Edge covering&\cellcolor{yellow!35}BS&HTO&3.66\%&2.74h&N/A&N/A&N/A&N/A\\
                5&Att-GCRN-GI&Edge covering&\cellcolor{yellow!35}w/o. HT&HTO&1.90\%&1.11h&2.62\%&1.21h&2.94\%&2.65h\\

			\hline
                6&Att-GCRN-GI&Edge covering&HTI&\cellcolor{yellow!35} no&9.80\%&0.18h&11.81\%&0.21h&14.70\%&0.52h\\
                7&Att-GCRN-GI&Edge covering&HTI&\cellcolor{yellow!35}LK&6.41\%&1.97h&N/A&N/A&N/A&N/A\\
                8&Att-GCRN-GI&Edge covering&HTI&\cellcolor{yellow!35}MCTS&3.42\%&1.19h&3.99\%&1.20h&N/A&N/A\\
                9&Att-GCRN-GI&Edge covering&HTI&\cellcolor{yellow!35}w/o. HT&3.31\%&1.14h&4.13\%&1.14h&5.42\%&2.62h\\  
                
			\Xhline{2\arrayrulewidth}

		\end{tabular}
		% \vspace{-4mm}
		\label{T4}
	\end{table*}
 \begin{figure*}[t!]
	\centering
    \setlength\abovedisplayskip{1pt}
\setlength\belowdisplayskip{1pt}
	\captionsetup[subfigure]{labelformat=empty}
		\subcaptionbox{\protect \label{P7}}[.33\textwidth]{
		\includegraphics[width=.29\textwidth]{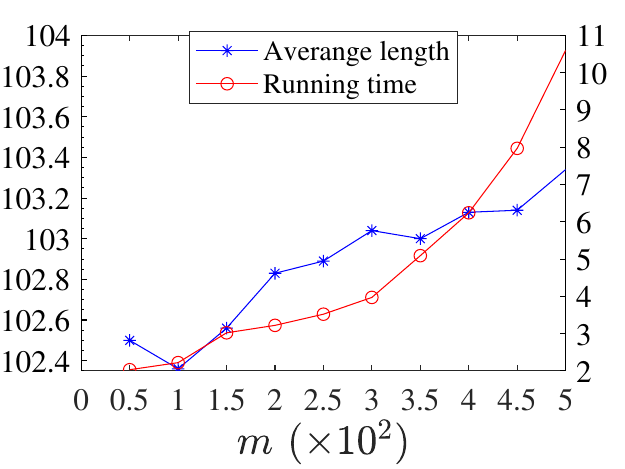}
	}
 \subcaptionbox{\protect \label{P8}}[.33\textwidth]{
		\includegraphics[width=.29\textwidth]{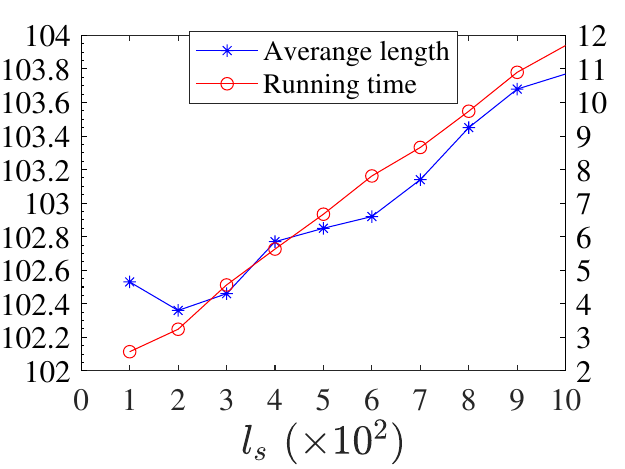}
	}
 \subcaptionbox{\protect \label{P9}}[.33\textwidth]{
    		\includegraphics[width=.29\textwidth]{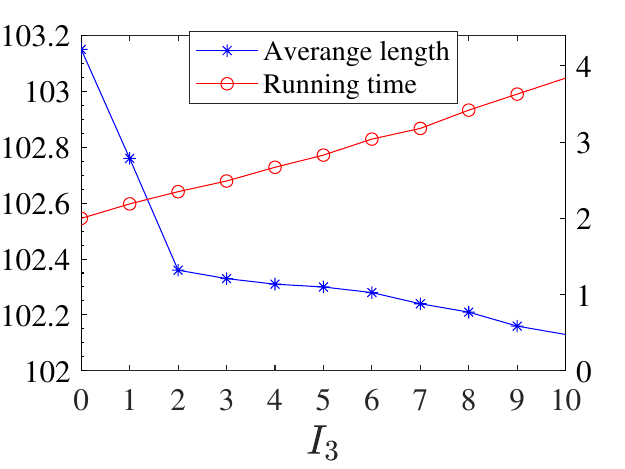}
    	}
	\vspace{-8.5mm}
	\caption{Parameter analysis.  In each subplot, the left and right y-axes show tour length and total running time, respectively.}
	\label{Para}
\end{figure*}

\stitle{Exp-5: Ablation Study.}
We further investigate the impact of different components in our model, including subgraph sampling, hyper tour, and hyper tour-guided initialization and optimization. The results are shown in Table \ref{T4}. 
First, we test the effectiveness of Att-GCRN-GI and replace it with Att-GCRN, shown as ID-1. Our Att-GCRN-GI outperforms Att-GCRN, which brings an average 0.10\% decrease in the average gap. Second, we test the effectiveness of
``edge covering" for subgraph sampling, and replace it with the ``node covering" from~\cite{fu2021generalize}, shown as ID-2. Our edge covering method outperforms the node covering method, which brings an average 0.25\% decrease in the average gap and 4.84\% decrease in running time on all datasets. 
Next, we compare the effects of our hyper tour-guided initialization (HTI) with different initialization methods, including the traditional ``greedy" and ``beam search (BS)" methods. We also test the results of randomly generating the order of supernodes without hyper tour guidance, denoted as ``w/o. HT". The results are shown as IDs-3 to 5.
Compared with the widely adopted greedy initialization method, our hyper tour-guided initialization reduces the average gap by 88.9\% , with total running times increasing from 3.21 to 4.06 hours. This increase in running time is acceptable given the significant reduction in the average gap. In contrast, the beam search-based initialization is substantially slower and could not be applied to TSP10K and TSP20K instances.
Besides, the introduction of the hyper tour on our tour initialization
reduces the average gap by 69.44\% and running time by 18.31\% across all datasets,
validating its effectiveness in improving convergence and overall results.
These findings justify the design of our hyper tour-guided initialization method, which improves convergence speed and leads to better overall results.

Finally, we compare our hyper tour-guided optimization (HTO) with other optimization methods, including the well-known ``LK" and ``MCTS" approaches, as well as the initial tour (no) and random edge selection for targeted neighborhood search, (``w/o HT"). {Our HTO reduces the average gap by 11.34\% over the initial solution, significantly outperforming other search methods.} Our HTO method outperforms MCTS, achieving significantly lower gaps (0.31\% vs. 3.42\% on TSP5K and 0.93\% vs. 3.99\% on TSP10K) and offers speed improvements of 21\% and 22.5\% on TSP5K and TSP10K, respectively. In contrast, LK performs worse than MCTS and is considerably slower. Both LK and MCTS fail to generalize to TSP20K, with LK unable to solve TSP10K.
Moreover, using MCTS with our initialization method outperforms AttGCRN+MCTS on both datasets, further demonstrating the superiority of our initialization approach. The introduction of the hyper tour in our optimization reduces the average gap by 82.27\% and running time by 17.14\% across all datasets. 
These results validate the effectiveness of our hyper tour-guided optimization in achieving high-quality final tours.

\stitle{Exp-6: Parameter Analysis.}
We tested the sensitivity of our method to key parameters, including the nearest neighbor number $m$, sub-tour length $l_{s}$, and the number of iterations for sub-tour optimization $I_3$. We varied $m$ from 50 to 500, $l_{s}$ from 100 to 1000, and $I_3$ from 0 to 10, while keeping other parameters consistent with Exp-1. 
The results are reported in Fig. \ref{Para}.

We can find that:
1) The average TSP length initially decreases with an increase in $m$ but then slightly increases, as shown in the left of Fig. \ref{Para}. This is because a larger range of optimization initially \vspace{-0.2mm}improves search results, but as $m$ continues to grow, the performance of LK search begins to deteriorate, affecting the targeted neighborhood search. The running time also increases rapidly with larger $m$; 
2) Similarly, the average TSP length decreases initially and then increases with the increase in $l_{s}$, as shown in the middle of Fig. \ref{Para}. Although the running time grows almost linearly with $l_{s}$, the dramatic runtime increase  does not sufficiently offset the decrease in tour length;
3) The average TSP length drops quickly at first and then more slowly with an increase in $I_3$, while the running time increases nearly linearly, as shown in the right of Fig. \ref{Para}. Therefore, we set $I_3$ to 2 to balance the solution quality and efficiency.

\section{Conclusion}\label{sec-con}
Combining neural-based methods has recently shown promise for large-scale TSPs, but they still face challenges in memory constraints, tour initialization and insufficient global guidance for efficient search. Our proposed Hyper Tour Guided Neighborhood Search (\mn~) method tackles these issues through a synergistic combination of sparse representation, hyper tour guidance, and targeted neighborhood search. By dividing and abstracting the TSP instance into supernodes and leveraging hyper tours as a global structural guide, our method simplifies the original large-scale problem into more manageable components, enabling more focused exploration and reducing redundant computations, which improves both initialization and optimization. This approach not only reduces the search space but also boosts solution efficiency and quality. Extensive experiments demonstrate that \mn~ represents the SOTA neural-based methods, successfully solving instances with up to 71,009 cities with a small optimality gap of just 3.68\%. In the future, we plan to parallelize our approach to handle even larger-scale TSP instances ($>$80,000 cities). Beyond the TSP, we also aim to extend the applicability of \mn~ to a broader range of combinatorial optimization problems, where efficiently solving very large-scale instances continues to pose significant challenges.

\bibliography{paper}
\bibliographystyle{icml2026}

%%%%%%%%%%%%%%%%%%%%%%%%%%%%%%%%%%%%%%%%%%%%%%%%%%%%%%%%%%%%%%%%%%%%%%%%%%%%%%%
%%%%%%%%%%%%%%%%%%%%%%%%%%%%%%%%%%%%%%%%%%%%%%%%%%%%%%%%%%%%%%%%%%%%%%%%%%%%%%%
% APPENDIX
%%%%%%%%%%%%%%%%%%%%%%%%%%%%%%%%%%%%%%%%%%%%%%%%%%%%%%%%%%%%%%%%%%%%%%%%%%%%%%%
%%%%%%%%%%%%%%%%%%%%%%%%%%%%%%%%%%%%%%%%%%%%%%%%%%%%%%%%%%%%%%%%%%%%%%%%%%%%%%%
\newpage
\appendix
\onecolumn
\appendix
\section{Supplemental Materials}
\label{sec-sup}
 \subsection{Details of Related Work} 
 \label{rel}
Traditional TSP solvers typically fall into three categories: exact methods~\cite{cook2011traveling}, approximation methods~\cite{dumitrescu2003approximation}, and heuristic methods~\cite{helsgaun2000effective}. With the rise of neural networks, recent research increasingly explores neural-based TSP methods or approaches that combine traditional solvers with neural networks. 
Early neural methods, typically limited to instances with up to 1,000 vertices, can be divided into constructive-based methods~\cite{vinyals2015pointer,bello2016neural,deudon2018learning,kool2018attention,khalil2017learning,li2024fast,grinsztajn2023winner,chalumeau2024combinatorial,luo2024neural,kwon2020pomo,jin2023pointerformer,drakulic2023bq,lyu2024scalable,jacobs21reinforcement,zhang2022learning,jiang2022learning,bi2022learning,zhou2023towards,son2023meta,li2025unify,wang2024asp} that generate tours from scratch with neural models and improvement-based methods~\cite{joshi2019efficient,xiao2024distilling,li2023distribution,min2024unsupervised,chen2019learning, wu2021learning,lu2019learning,hudson2022graph,yao22reversible,ma2023learning,ye2023deepaco,wang2025efficient,li2025unify} that first generated a tour and then iteratively refined under neural guidance toward better solutions.
\stitle{Constructive-based Methods.}
These methods generate TSP tours from scratch using neural models. The Pointer Network~\cite{vinyals2015pointer} first introduced an encoder-decoder framework for supervised incremental solution generation. Later advancements incorporated reinforcement learning~\cite{bello2016neural,deudon2018learning,kool2018attention,khalil2017learning}, diffusion models~\cite{li2024fast}, refined decoders~\cite{grinsztajn2023winner,chalumeau2024combinatorial,luo2024neural}, utilized TSP tour symmetry~\cite{kwon2020pomo,jin2023pointerformer,drakulic2023bq} and the nearest neighbors~\cite{lyu2024scalable}, trained on diverse instance distributions~\cite{jacobs21reinforcement,zhang2022learning,jiang2022learning,bi2022learning}, explored meta-learning~\cite{zhou2023towards,son2023meta}, joint probability estimation~\cite{li2025unify} and  curriculum learning~\cite{wang2024asp}. However, these approaches are limited to TSP instances with up to 1,000 vertices and struggle with larger instances.

\stitle{Improvement-based Methods.} 
These methods combine neural models with traditional search methods, and can be classified into two types. The first type is to generate the heatmap matrix (probability of an edge being in the optimal solution) based on the GNN model and then use search methods like {beam search~\cite{joshi2019efficient,xiao2024distilling}, 2-opt~\cite{li2023distribution} and best-first local search~\cite{min2024unsupervised} to optimize the solution.
% like  to optimize the solution~\cite{joshi2019efficient, li2023distribution, min2024unsupervised,xiao2024distilling}. 
Another type of search-based methods adopts neural models (mainly reinforcement learning) to guide the local search, including 2-opt~\cite{chen2019learning, wu2021learning,lu2019learning,hudson2022graph,yao22reversible}, k-opt~\cite{ma2023learning}, beam search~\cite{choo2022simulation} and Ant Colony Optimization~\cite{ye2023deepaco}.}
The lack of optimal labels for large-scale instances makes training neural network models challenging, which restricts these methods to handling instances with up to 1,000 vertices.

\subsection{Att-GCRN with Geometric Information.}\label{attgcrn}
The main structure of our Att-GCRN-GI is consistent with Att-GCRN, with the only difference being in the input layer. Therefore, in this part, we first introduce the pretrained encoder-decoder framework for the geometric information, and then  Att-GCRN with geometric information.

\subsubsection{Pretrained Encoder-Decoder Framework for the Geometric Information}
Inspired by the work of \cite{zhou2023improving}, we construct a pretrained encoder-decoder network and incorporate the geometric information-aware hidden representations from the encoder as part of the node features for Att-GCRN. 
Specifically, our model uses an encoder that comprises an MLP layer to map the coordinates of vertices in $\mathbb{V}$ into a high-dimensional vector space, followed by layer normalization. This process can be formalized as follows:
\begin{equation} 
\setlength\abovedisplayskip{2pt}
\setlength\belowdisplayskip{2pt}
    \mathcal{F} = \mathrm{LayerNorm}( \mathcal{M} (  \mathcal{V} ) )   
\end{equation}
where $\mathcal{M}(\cdot)$ refers to the MLP layer and $\mathcal{V}$ is the feature matrix composed of the coordinates of all vertices in $\mathbb{G(V, E)}$.

The decoder is designed to approximate the high-order geometric information of each triplet set, such as the edge lengths and triangle areas. 
To integrate geometric encodings with the vertex features of Att-GCRN,  we first employ a triangle sampler module to randomly sample multiple triplet sets, denoted as $\mathbb{V}_s = \{v_a, v_b, v_c\}$, from the vertex set $\mathbb{V}$. For each triplet set $\mathbb{V}_s$, we derive their corresponding features $\mathcal{F}_s = [f_a, f_b, f_c] \subset \mathcal{F}$ produced by the encoder, then process them through $l_n$ attention layers, and output the intermediate features $\hat{\mathcal{F}}_s = \{ \hat{f}_a, \hat{f}_b, \hat{f}_c \}$. Subsequently, two MLPs, $\mathcal{G}_d$ and $\mathcal{G}_a$, are used to predict the edge lengths and triangle areas, respectively, which can be formalized as: 
\begin{equation}
\setlength\abovedisplayskip{1pt}%shrink space
\setlength\belowdisplayskip{1pt}
\begin{aligned}
    \mathcal{D} & = [\mathcal{G}_d(\hat{f}_a || \hat{f}_b), \mathcal{G}_d(\hat{f}_a || \hat{f}_c), \mathcal{G}_d(\hat{f}_b || \hat{f}_c)] \\
    \mathcal{A} & = \mathcal{G}_a(\hat{f}_a || \hat{f}_b || \hat{f}_c)
\end{aligned}
\end{equation}
where $||$ is the concatenation operator.
We use the mean squared error (MSE) between predicted geometric information $\mathcal{D}$ and $\mathcal{A}$, and the ground-truth geometric information $\hat{\mathcal{D}}$ and $\hat{\mathcal{A}}$ of the sampled triplet sets as its loss function.
\begin{equation}
\setlength\abovedisplayskip{1pt}%shrink space
\setlength\belowdisplayskip{1pt}
\begin{aligned}
        \mathcal{J}_\texttt{GI} = \sum_{i=1}^{T} \big( \sum_{j=0}^2\mathrm{MSE}(\mathcal{D}_{i,j}, \hat{\mathcal{D}}_{i,j}) + \mathrm{MSE}(\mathcal{A}_i, \hat{\mathcal{A}}_i) \big)
\end{aligned}
\end{equation}
where $\mathcal{D}_{i,j}$ is the $j$-th edge's length of the $i$-th triple set, and $T$ is the number of sampled triplet sets.

We incorporate the encoder's hidden representations $\mathcal{F}$ with the coordinate features to form a new vertex feature for the Att-GCRN.
\begin{equation}
    \setlength\abovedisplayskip{2pt}%shrink space
\setlength\belowdisplayskip{2pt}
\hat{\mathcal{V}} = \mathcal{V} || \mathcal{F}
\end{equation}

\subsubsection{Att-GCRN-GI}
We next introduce the Att-GCRN with the geometric information.

\stitle{Input layer.} The Att-GCRN model takes the incorporated feature 
 $\hat{\mathcal{V}}$ as input and embeds it to H dimension features, which is shown in Equ.~\ref{node_embedding}. 
\begin{equation}
    \begin{aligned}
        % \hat{F}_i &= A_1x_i + b_1\\
        \mathcal{V}^0 &= A_1 \cdot \hat{\mathcal{V}}\label{node_embedding}
    \end{aligned}
\end{equation}
where $A_1 \in \mathbb{R}^{H\times 2}$.
The edge distance $d_{i,j}$ is embedded as a $\frac{H}{2}$ dimensional feature vector. Then it defines an indicator function of a TSP edge $\delta_{i,j}^{K-NN}$ with the value one if nodes $i$ and $j$ are
K-nearest neighbors, value two for self-connections, and value zero otherwise. Thus the edge input feature $e_{i,j}$ is:
$$e_{i,j}^0 = A_2d_{i,j} + b_2||A_3\delta_{i,j}^{k-NN}$$
where $A_2 \in \mathbb{R}^{H/2\times 1}$ ,$A_3 \in \mathbb{R}^{H/2\times 3}$. The introduction of the K-nearest neighbor could help accelerate the learning process.
 
\stitle{Graph Convolution Layer.} Let $\mathcal{V}_i^l$ and $e_{i,j}^l$ denote respectively the node feature vector and edge feature vector at layer $l$ associated with node $i$ and edge $(i,j)$. The node and edge features at the next layer is:
\[
\setlength\abovedisplayskip{2pt}%shrink space
\setlength\belowdisplayskip{2pt}
\mathcal{V}_i^{l+1} = \mathcal{V}_i^{l} + ReLU(BN(W_1^l\mathcal{V}_i^{l} + \sum_{j~i}\eta_{i,j}^l \odot W_2\mathcal{V}_j^{l})),\]
\[\setlength\abovedisplayskip{2pt}%shrink space
\setlength\belowdisplayskip{2pt}
\mathrm{with} \  \eta_{i,j}^l = \frac{\sigma(e_{i,j}^l)}{\sum_{j'~i}\sigma(e_{i,j'}^l) + \epsilon}\]
\[
\setlength\abovedisplayskip{2pt}%shrink space
\setlength\belowdisplayskip{2pt}
e_{i,j}^{l+1} = e_{i,j}^{l} + ReLU(BN(W_3^le_{i,j}^{l} + W_4^l\mathcal{V}_i^{l} + W_5^l\mathcal{V}_j^{l}))\]
where $W_1, W_2, W_3, W_4\  \mathrm{and} \  W_5 \in \mathbb{R}^{H\times H}$, $\sigma$ is the sigmoid function, $\epsilon$ is a small value, $ReLU$ is the rectified linear unit, and $BN$ stands for batch normalization.

\stitle{MLP Classifier.} The edge embedding of the last layer is used to generate the heatmap through an MLP layer as follows. 
\[\mathrm{heatmap}_{i,j} = MLP_h(e^L_{i,j})\] 

\stitle{Loss Function.} Given the ground-truth tour, we could convert it into an adjacency matrix $\mathrm{heatmap}^{GT}$ and minimize the binary cross-entropy of the $\mathrm{heatmap}$ and $\mathrm{heatmap}^{GT}$.

\subsection{Time Complexity Analysis}\label{complexity}
Our method comprises five key components: sparse heatmap graph construction, hyper tour generation, initialization, targeted neighborhood search, and sub-tour optimization, with GCRN model and LK search employed to address various subproblems. The complexity of GCRN model is $O(n^2)$~\cite{Wu21Comprehensive}, while LK search is a PLS-complete problem~\cite{papadimitriou1992complexity}, and its complexity cannot be precisely defined but is denoted as $O(\texttt{L}(n))$.

Our sparse heatmap graph construction involves solving the set covering problem using a greedy algorithm, which has a time complexity of $O(cn)$. Additionally, generating sub-heatmaps with the GCRN model takes $O(cp^2)$ time, where $n$, $c$, and $p$ represent the number of vertices, candidate sets, and subgraphs, respectively\footnote{In our method, both $c$ and $p$ are significantly smaller than $n$. For an instance with 10,000 points, the average number of candidate sets $c$ is 368, with $p$ fixed at 100.}. Furthermore, merging sub-heatmaps requires traversing all edges of each subgraph, which also has a time complexity of $O(cp^2)$. Consequently, the overall time complexity is $O(c(n + p^2))$.

For hyper tour generation, the process of obtaining connected subgraphs and deleting bridges has a complexity of $O((k+1)n)$, given that the number of edges in $G_s$ is less than $kn$. Consequently, the clustering process, which involves iteratively extracting subgraphs and deleting bridges, takes $O(I_1(k+1)n)$ time, where $I_1$ represents the number of iterations needed to obtain connected subgraphs from $G_s$. The function $\texttt{solve\_small\_tsp}$, which integrates GCRN and LK search, contributes $O(n^2 + L(n))$ to the overall complexity. Therefore, the total time complexity for hyper tour generation phase is $O(n^2 + L(n) + I_1(k+1)n)$.

The tour initialization phase employs the LK search on sub-tours of length $l_s$ and is repeated $n/l_s$ times, resulting in a time complexity of $O(nL(l_s)/l_s)$.
In the targeted neighborhood search process, identifying the longest edge among candidate edges and forming subproblems takes $O(n)$, while solving each subproblem contributes $O(L(m))$ to the overall complexity. Consequently, the total time complexity for the targeted neighborhood search phase is $O(I_2(n + L(m)))$, where $I_2$ represents the number of iterations and $m$ denotes the number of neighbors.
Finally, the sub-tour optimization step has a time complexity of $O(I_3 \times (nL(l_s)/l_s))$, where $I_3$ is the number of iterations and $l_s$ is the length of the sub-tours.

\subsection{Impact of Hyper Tour Quality on Final Solution Performance}\label{exp:hyper}

To empirically investigate the interaction between hyper tour quality and final solution performance, we design a controlled experiment by injecting varying levels of noise into the heatmap during its construction. This produces 100 different hyper tours on the same instance, each leading to a different initial solution. We use the length of the initial solution (before optimization) as a proxy for hyper tour quality—the shorter the initial solution, the better the hyper tour. We then apply the same optimization procedure to each initial solution and group the results into 10 bins based on initial tour length. For each bin, we report the number of instances, the average initial solution length, the average final optimized length, and the average solving time.

\begin{table}[h]
\centering
\caption{Impact of initial hyper tour quality on final solution performance.}
\resizebox{0.46\textwidth}{!}{
\begin{tabular}{c c c c c c}
\hline
\multirow{2}{*}{Bin ID} & Initial Tour & \multirow{2}{*}{\# Instances} & Avg. Initial  & Avg. Final  & Avg. Solving  \\
& Length Range&&Length&Length&Time (h)\\
\hline
1  & 116--117 & 27 & 116.3 & 102.1 & 0.22 \\
2  & 117--118 & 19 & 117.4 & 102.1 & 0.23 \\
3  & 118--119 & 13 & 118.6 & 102.1 & 0.24 \\
4  & 119--120 & 10 & 119.3 & 102.1 & 0.25 \\
5  & 120--121 & 5  & 120.6 & 102.1 & 0.28 \\
6  & 121--122 & 6  & 121.4 & 102.1 & 0.31 \\
7  & 122--123 & 8  & 122.7 & 102.3 & 0.33 \\
8  & 123--124 & 5  & 123.5 & 102.4 & 0.38 \\
9  & 124--125 & 4  & 124.6 & 102.7 & 0.40 \\
10 & 125--126 & 3  & 125.3 & 103.1 & 0.41 \\
\hline
\end{tabular}}
\label{tab:hyper_quality}
\end{table}

As shown in Table~\ref{tab:hyper_quality}, the final solution quality remains largely stable as long as the hyper tour quality stays above a certain threshold (Bins 1--6). Beyond this point, as the hyper tour quality deteriorates further, the final solution quality begins to degrade slightly and the optimization time increases. These results demonstrate that while high-quality hyper tours help reduce solving time, our optimization procedure is robust and consistently capable of refining even suboptimal initial solutions.

\subsection{Discussion on the Adoption of the LK Search}\label{sup-LK}
Lin-Kernighan (LK) search is a powerful local search heuristic designed for solving small scale the Traveling Salesman Problem (TSP). It is an extension of the k-opt heuristic, dynamically selecting the most promising edge exchanges rather than fixing k in advance. By adaptively exploring multiple swap possibilities, LK search efficiently escapes local optima and improves solution quality. 

Current heuristic and neural-base methods typically rely on LK search as the foundational unit in their local search due to its robust performance and efficiency on smaller TSP instances, such as the current SOTA method LKH, MCTS based neural methods. Following these methods, we choose to use LK search as the fundamental unit of our search. We also experimented with other simple search methods such as 2-opt, 3-opt, or-opt, and k-opt, but their performance was far inferior to LK search. We simply chose the most commonly used one. The results for 2-opt, 3-opt, or-opt, and k-opt were also good and exceeded the baseline, but were inferior to LK search.

%%%%%%%%%%%%%%%%%%%%%%%%%%%%%%%%%%%%%%%%%%%%%%%%%%%%%%%%%%%%%%%%%%%%%%%%%%%%%%%
%%%%%%%%%%%%%%%%%%%%%%%%%%%%%%%%%%%%%%%%%%%%%%%%%%%%%%%%%%%%%%%%%%%%%%%%%%%%%%%

\end{document}